\documentclass[a4paper,fleqn,numbers]{cas-sc}

\usepackage[numbers,sort&compress]{natbib}
\usepackage{graphicx}
\usepackage{subcaption}

\begin{document}
\let\WriteBookmarks\relax
\def\floatpagepagefraction{1}
\def\textpagefraction{.001}

\shorttitle{TeachPro}

\shortauthors{Xiangqian Wang et~al.}

\title [mode = title]{TeachPro: Multi-Label Qualitative Teaching Evaluation via Cross-View Graph Synergy and Semantic Anchored Evidence Encoding}

\author[1]{Xiangqian Wang}
\ead{wxqpds@126.com}
\credit{Conceptualization; Funding acquisition; Resources; Supervision; Writing - original draft}

\author[2]{Yifan Jia}
\ead{jiayf@hrbeu.edu.cn}
\credit{Visualization; Writing - review \& editing}

\author[3]{Yang Xiang}
\ead{3120230912417@stu.xhu.edu.cn}
\credit{Software; Validation; Writing - original draft}

\author[4]{Yumin Zhang}
\cormark[2]
\ead{24241214904@stu.xidian.edu.cn}
\credit{Validation; Visualization; Writing - review \& editing}

\author[5]{Yanbin Wang}
\cormark[1]
\ead{wangyanbin15@mails.ucas.ac.cn}
\credit{Methodology; Formal analysis; Investigation; Supervision; Writing - review \& editing}

\author[1]{Ke Liu}
\ead{905854383@qq.com}
\credit{Data curation; Project administration; Software; Supervision}

\affiliation[1]{organization={Pingdingshan University},
    addressline={School of Information Engineering},
    city={Pingdingshan},
    country={China}}

\affiliation[2]{organization={Harbin Engineering University},
    addressline={Yantai Research Institute},
    city={Yantai},
    country={China}}

\affiliation[3]{organization={Xihua University},
    addressline={School of Computer and Software Engineering},
    city={Chengdu},
    country={China}}

\affiliation[4]{organization={Xidian University},
    addressline={Hangzhou Institute of Technology},
    city={Hangzhou},
    country={China}}

\affiliation[5]{organization={Shenzhen MSU-BIT University},
    addressline={Department of Electronics and Computer Engineering},
    city={Shenzhen},
    country={China}}

\cortext[1]{Corresponding author}
\cortext[2]{Co-corresponding author}

\begin{abstract}
Standardized Student Evaluation of Teaching (SET) often suffer from low reliability, restricted response options, and response distortion. Existing machine learning methods that mine open-ended comments usually reduce feedback to binary sentiment, which overlooks concrete concerns such as content clarity, feedback timeliness, and instructor demeanor, and provides limited guidance for instructional improvement.

We propose TeachPro, a multi-label learning framework that systematically assesses five key teaching dimensions: professional expertise, instructional behavior, pedagogical efficacy, classroom experience, and other performance metrics. We first propose a Dimension-Anchored Evidence Encoder, which integrates three core components: (i) a pre-trained text encoder that transforms qualitative feedback annotations into contextualized embeddings; (ii) a prompt module that represents five teaching dimensions as learnable semantic anchors; and (iii) a cross-attention mechanism that aligns evidence with pedagogical dimensions within a structured semantic space. We then propose a Cross-View Graph Synergy Network to represent student comments. This network comprises two components: (i) a Syntactic Branch (SynGCN) that extracts explicit grammatical dependencies from parse trees, and (ii) a Semantic Branch (SemGCN) that models latent conceptual relations derived from BERT-based similarity graphs. To harmonize distinct views, a bi-affine fusion module aligns syntactic and semantic units, while a differential regularizer enforces embedding disentanglement, promoting complementary representations. Finally, a cross-attention mechanism bridges the dimension-anchored evidence with the multi-view comment representations. We also contribute a novel benchmark dataset featuring expert qualitative annotations and multi-label scores. Extensive experiments demonstrate that TeachPro offers superior diagnostic granularity and robustness across diverse evaluation settings. Code and dataset are available at: https://github.com/gmmmmod2/QDTL.
\end{abstract}


\begin{highlights}
\item TeachPro performs multi-label SET across five actionable teaching dimensions.
\item Evidence encoder aligns comments with learnable dimension anchors.
\item Dual-branch GCNs model syntactic dependencies and semantic similarity relations.
\item Bi-affine fusion and differential regularization ensure complementary views.
\item New expert-annotated benchmark dataset; TeachPro shows robust, granular gains.
\end{highlights}


\maketitle

\section{Introduction}

Student Evaluation of Teaching (SET) is central to quality assurance and evidence-informed decision making in higher education \cite{Zhao2022Literature}. Historically, institutions have relied on structured scales composed of closed-ended items aligned with canonical dimensions such as clarity, organization, and feedback. While such instruments afford standardization and comparability, they are vulnerable to halo effects, response biases, and a lack of contextual depth, which can obscure the multi-faceted nature of instructional quality \cite{Shevlin01122000}. In parallel, universities increasingly collect open-ended comments that articulate students’ experiences in their own words, providing richer depictions of classroom practice and course design \cite{Stupans2016}. The practical use of this textual information at scale, however, requires automated analysis beyond manual coding, motivating the adoption of Natural Language Processing (NLP) techniques \cite{Kastrati2021survey,Shaik2023survey}.

Existing sentiment-based approaches have demonstrated potential in processing student feedback by deriving overall polarity classifications from comments \cite{Okoye2020,Okoye2022}. However, critical limitations persist in achieving meaningful, actionable instructional evaluation.
\begin{enumerate}
    \item \textbf{Over-simplified evaluation:} Prevailing methods collapse student comments into a single sentiment label, failing to capture heterogeneous opinions across different teaching aspects. For instance, a comment such as "The lectures are incredibly clear and well-organized, but the feedback on assignments is often delayed and not very helpful" would be flattened into a single, ambiguous sentiment score. This completely obscures the nuanced reality: strong positivity regarding the instructor's "content clarity" and "instructional conduct", coupled with strong negativity concerning "feedback timeliness" and "supportiveness".
    \item \textbf{Lack of evidence grounding:} Current models process student comments in isolation, disregarding both qualitative evidence annotations and structured mechanisms for associating loosely-organized semantic content with pedagogical quality indicators. For example, without explicit guidance, a model may struggle to link a student's phrase "goes off on tangents" to the formal teaching dimension of "Instructional Conduct", or to recognize that "can't hear clearly in the back" constitutes critical evidence for evaluating "Classroom Experience". The lack of a framework to ground such informal, open-ended utterances into a structured assessment matrix represents a fundamental gap.
    \item \textbf{Neglect of educational linguistic: } These approaches treat educational feedback as generic text, neglecting to model the distinctive syntactic structures and conceptual relationships characteristic of student discourse. For example, the syntactic negation in "I wouldn't say the lectures are uninteresting" can completely reverse the surface-level sentiment, or fail to connect conceptually related phrases like "tough grader" and "high standards" Generic text classifiers fail to explicitly model these crucial syntactic and conceptual cues.
    \item \textbf{Limited datasets:} Available public datasets remain limited to comment-sentiment pairs, lacking the multi-dimensional, fine-grained annotations necessary for comprehensive teaching assessment.
\end{enumerate}

To address these limitations, we reformulate the Student Teaching Evaluation (SET) task as a structured multi-dimensional multi-classification problem and propose TeachPro, a Multi-aspect Teaching Profiler that systematically bridges unstructured student feedback with pedagogical assessment. Our approach begins with a Dimension-Anchored Evidence Encoder based on prompt learning, where each evaluation dimension is paired with learnable prompts dynamically refined through cross-attention with dimension-specific evidence snippets extracted from comments, thereby establishing explicit grounding between textual evidence and pedagogical concepts. It then constructs syntactic and semantic graphs from raw comments, performing concurrent message passing to capture both formal linguistic structures and contextual conceptual relationships, enabling robust representation learning tailored to educational discourse. Finally, it employs a low-rank adaptation architecture that maintains a shared base classifier while applying dimension-specific transformations via lightweight rank-constrained matrices, effectively balancing model efficiency and discriminative capacity across teaching aspects. The entire framework is supported by our newly constructed benchmark dataset, which includes multi-dimensional fine-grained annotations (five teaching dimensions, each with ternary classifications), providing essential supervision for evidence-aware, dimension-specific learning. Collectively, these components deliver a comprehensive, interpretable, and computationally efficient solution that transforms free-form student comments into structured pedagogical assessments.

Our key contributions are as follows:

\begin{itemize}

\item \textbf{Task formulation.} This research fundamentally reframes educational feedback analysis from the prevalent paradigm of binary sentiment classification to a structured framework of discrete evaluations across multiple pedagogical dimensions. Our proposed method, TeachPro, successfully instantiates this unprecedented paradigm, demonstrating robust predictive performance and fine-grained disaggregation capability across diverse evaluation settings.
\item \textbf{Evidence Encoder.} We propose a dimension-anchored evidence encoder that transforms qualitative instructional annotations into contextualized evidence embeddings, represents teaching evaluation aspects as learnable semantic anchors, and achieves semantic association between annotated evidence and structured pedagogical dimensions via cross-attention mechanisms.
\item \textbf{Student Comment Encoder.} We propose a Cross-View Graph Synergy Network for student comment encoding that uses parallel graph convolutional pathways to learn syntactic and semantic representations capturing dependencies and conceptual relationships, aligns these via a bi-affine fusion module, and yields unified embeddings that capture linguistic structure and conceptual semantics.

\item \textbf{Parameter-efficient prediction.} We introduce a low-rank adaptation (LoRA) classification head that maintains parameter efficiency through a shared global projection while enabling dimension-specific discrimination via lightweight, rank-constrained transformations. This design achieves an optimal balance between model compactness and representational capacity for multi-dimensional assessment tasks.

\item \textbf{Dataset.} We contribute a benchmark dataset that features open-ended student comments annotated through a multi-tiered framework: 1) multi-label mapping across five pedagogical dimensions, 2) qualitative evidence annotations linking specific comment segments to corresponding dimensions, and 3) fine-grained ternary classification (good/fair/poor) for each dimension. This structured annotation provides the first  resource supporting evidence-based interpretation and granular educational feedback across multiple teaching evaluation facets.
\end{itemize}

\section{Related Work}
Early Student Evaluation of Teaching (SET) research largely relied on closed-ended, scale-based instruments. Spooren \cite{Spooren01122007} designed a ten–Likert-scale instrument aligned with predefined dimensions (e.g., clarity, engagement, content delivery) to address large-scale benchmarking and cross-course comparability in higher education. Extending this line, Spooren \cite{Spooren2013} conducted a state-of-the-art synthesis to evaluate the validity of scale-based SET, highlighting where such instruments succeed and where they fall short. More recently, Cannon and Cipriani \cite{Cannon2021} designed a quantitative framework to measure and \emph{quantify} halo effects in SET, and Michela \cite{Michela2022} offered a reanalysis and alternative interpretation to quantify halo effects and clarify how global impressions bias multiple rating dimensions. While these designs enable institution-level monitoring and standardized aggregation, they remain vulnerable to halo-induced distortions that undermine diagnostic precision and validity across dimensions \cite{Cannon2021, Michela2022}; moreover, their score-centric format provides limited explanatory depth about instructional practices \cite{Spooren2013}.

To capture richer, context-sensitive teaching signals, later work pivots to narrative comments. Fuller \cite{Fuller2024} designed a practical workflow that explores using ChatGPT \cite{brown2020language} to analyze student course evaluation comments, aiming to surface qualitative insights that numeric scales miss. In parallel, Sunar and Khalid \cite{SunarKhalid2024} designed a systematic review protocol to map the landscape of NLP for student feedback, identifying typical pipelines, data practices, and open challenges for educational settings. Although these efforts show the value of free-text analysis, manual interpretation does not scale institution-wide \cite{SunarKhalid2024}, and naïve automation can miss fine-grained, pedagogically meaningful cues present within the same comment \cite{Fuller2024}.

With scalability in view, many studies adopt NLP-based sentiment analysis and opinion mining. Wankhade \cite{Wankhade2022} designed a comprehensive survey that systematizes sentiment-analysis methods and application patterns, offering guidance for educational deployments. Building a task-specific pipeline, Ren \cite{ren2023automatic} designed an aspect-level automatic scoring approach for student feedback using LSTM/BERT-style backbones to address the need for rapid polarity detection and preliminary aspect cues in course evaluations. For modeling choices, Tzimiris \cite{Tzimiris2025TopicSA} designed a comparative evaluation of transformer-based language models for topic-based sentiment analysis, informing architecture selection and performance trade-offs for large-scale opinion mining. Most pipelines still formulate comment-level sentiment as coarse three-way polarity, collapsing heterogeneous sentiments within a single comment (e.g., praise for clarity but critique of feedback timeliness) into one global label; this reduces interpretability and actionability and typically lacks explicit mechanisms to disentangle aspect-specific sentiment patterns \cite{Hua2024SLR}.

Closer to pedagogical needs, Aspect-Based Sentiment Analysis (ABSA) targets dimension-level signals. Alotaibi \cite{Alotaibi2024ABSA} designed an ABSA framework for open-ended responses in student course evaluation surveys to attach aspect-aligned sentiment to instructional dimensions, while Sindhu \cite{sindhu2019aspect} designed an aspect-based opinion-mining pipeline for faculty performance evaluation to address fine-grained judgments beyond overall polarity. To support research infrastructure, Herath \cite{Herath2022Dataset} designed a dataset and baseline for automatic student feedback analysis, facilitating reproducible benchmarking of ABSA-style models in education. Complementarily, Hua \cite{Hua2024SLR} designed a systematic review of ABSA that consolidates modeling patterns—from sequence tagging and attention to transformer encoders—and surfaces gaps relevant to education-oriented deployments. Existing educational ABSA efforts often fix aspect label sets without modeling inter-aspect interactions; few incorporate prototype-driven attention or prompt-based semantic alignment with pedagogical dimensions, and, to our knowledge, no work leverages prompt-generated embeddings tailored to teaching-evaluation facets. Moreover, the scarcity of publicly available datasets annotated with structured, dimension-level sentiment labels \cite{Herath2022Dataset} continues to limit progress toward fine-grained, interpretable, and pedagogically meaningful SET analysis.

\begin{figure}
    \centering
    \includegraphics[width=0.4\textwidth]{ 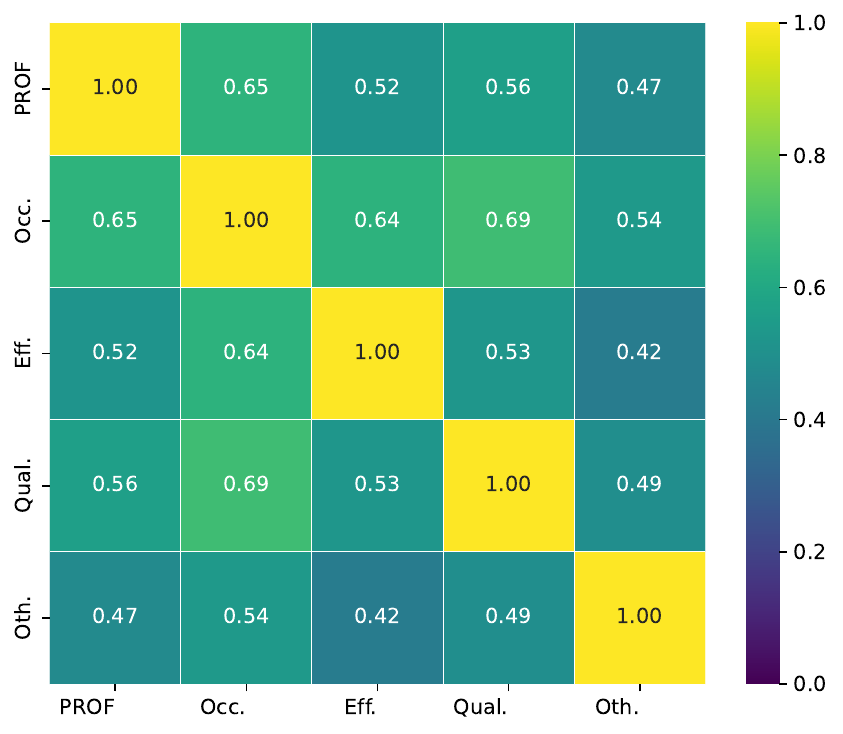} 
    \caption{Heatmap of correlation between different dimensions in the dataset}
    \label{fig:1pdf}
\end{figure}

\section{Datasets}
We introduce TeachScope, a pedagogically-grounded benchmark dataset for teaching evaluation, featuring multi-dimensional quality ratings coupled with evidence-based textual justifications, constructed through systematic manual annotation of student comments from publicly available professor reviews \cite{he2020big}. 

Our annotation protocol operationalizes teaching quality through five pedagogically-grounded dimensions:
\begin{enumerate}
    \item Professional Expertise: Subject matter mastery and knowledge depth.
    \item Instructional Conduct: Teaching attitude, professionalism, and responsibility.
    \item Teaching Effectiveness: Impact on learning outcomes and grade achievement.
    \item Classroom Experience: Overall teaching quality and classroom atmosphere.
    \item Other Performance: Additional teaching-related aspects.
\end{enumerate}

The annotation process involved a three-stage pipeline:
\begin{itemize}
    \item Removed blank and semantically empty comments from the raw corpus.
    \item  Two trained annotators independently scored each comment across all five dimensions. Each evaluation dimension is assessed using a 3-point ordinal scale (0=Negative, 1=Neutral/Not Mentioned, 2=Positive).
    \item For each dimension score, annotators identified specific text spans from the original comment that justified the assigned rating.
\end{itemize}

The annotation process emphasized the following core protocols: 
\textbf{Direct Quotation Principle}: Every dimensional score must be directly justified by specific phrases extracted from the comment text, grounding the assessment in tangible evidence. 
\textbf{Independence Principle}: Scores for different dimensions must be based on distinct evidence snippets, preventing a single phrase from unduly influencing multiple, unrelated facets of the evaluation. 
\textbf{Negative Evidence Handling}: A clear distinction was enforced between the absence of mention (which should not be penalized) and the presence of explicit negative feedback (which must be scored accordingly).

A multi-layered quality control system was implemented to ensure annotator consistency and data reliability: 
\textbf{Calibration Test}: Prospective annotators were required to pass a qualification test comprising 20 gold-standard samples, achieving an inter-annotator agreement ($Kappa \geq 0.95$) before proceeding to the full dataset. 
\textbf{Dynamic Validation}: After every 50 annotations, annotators re-calibrated their judgment by scoring a set of predefined standard samples to prevent concept drift over time. 
\textbf{Adjudication Process}: All scoring discrepancies between the two primary annotators were resolved through arbitration by a senior annotator, establishing a final ground-truth label.

The final dataset comprises student comments paired with five-dimensional ordinal ratings and corresponding evidentiary annotations. An overview of the entire dataset is provided in Table \ref{tab:schema}. Additionally, a correlation heatmap (shown in Figure \ref{fig:1pdf}) was generated to visualize statistical correlations among variables across these dimensions.

\begin{table}[width=\linewidth,cols=4,pos=ht]
\caption{TeachScope Dataset Structure and Annotation Schema. Each record contains a student comment and five evaluation dimensions. For each dimension, annotators assign an ordinal score (0=Negative, 1=Neutral/Not Mentioned, 2=Positive) and extract supporting text spans.}
\label{tab:schema}
\begin{tabular*}{\tblwidth}{@{} L L L C @{}}
\toprule
\textbf{Field Name} & \textbf{Type} & \textbf{Description} & \textbf{Label Range} \\
\midrule
\texttt{professor\_name} & Categorical & Identifier of the evaluated instructor. & -- \\
\texttt{comments}        & Text        & Original student review describing teaching experience. & -- \\
\midrule
\texttt{Professionalism\_score}  & Ordinal   & Measures subject-matter expertise and academic professionalism. & 0--2 \\
\texttt{Professionalism\_reason} & Text span & Evidence phrase supporting the professionalism score. & -- \\
\texttt{Occupational\_score}     & Ordinal   & Reflects teaching attitude, fairness, and responsibility. & 0--2 \\
\texttt{Occupational\_reason}    & Text span & Phrase indicating attitude or behavioral cues. & -- \\
\texttt{Effectiveness\_score}    & Ordinal   & Captures learning outcomes and instructional clarity. & 0--2 \\
\texttt{Effectiveness\_reason}   & Text span & Sentence describing course difficulty or outcomes. & -- \\
\texttt{Quality\_score}          & Ordinal   & Represents classroom experience and overall teaching quality. & 0--2 \\
\texttt{Quality\_reason}         & Text span & Comment referring to class atmosphere or engagement. & -- \\
\texttt{Other\_score}            & Ordinal   & Miscellaneous aspects beyond the primary dimensions. & 0--2 \\
\texttt{Other\_reason}           & Text span & Additional evidence for other aspects. & -- \\
\bottomrule
\end{tabular*}
\end{table}

\section{Methodology}
\begin{figure*}
    \centering
    \includegraphics[width=1\textwidth]{ 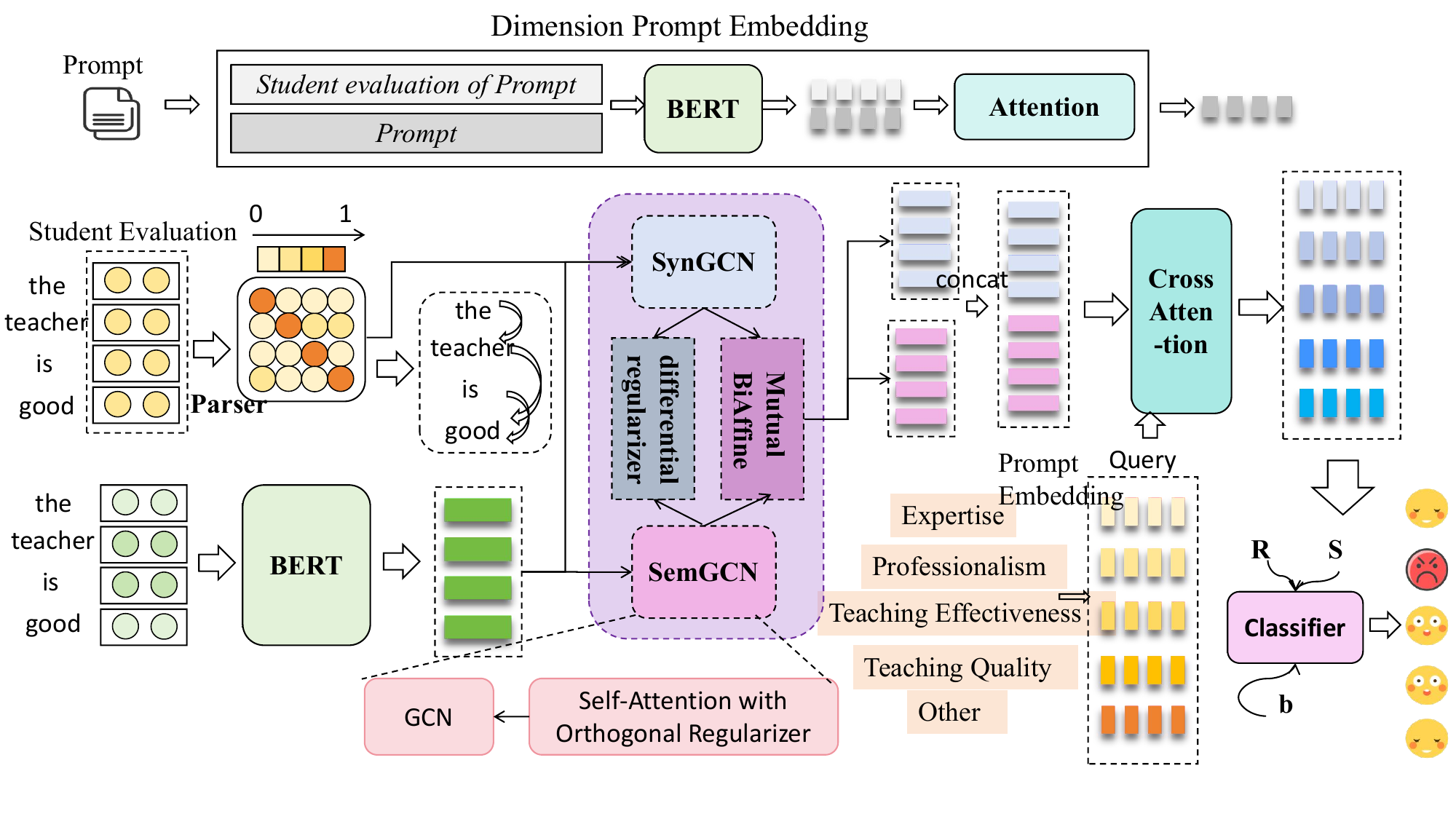} 
    \caption{The overall framework of the model in this article.}
    \label{fig:model} 
\end{figure*}

We propose TeachPro, a deep learning architecture designed for fine-grained, multi-dimensional teacher evaluation. As illustrated in Fig.~\ref{fig:model}, the framework comprises three core components:
\begin{itemize}
    \item Cross-View Graph Synergy Network: This module initializes comment representations using a pre-trained BERT encoder and constructs parallel syntactic and semantic graphs. It captures structural and relational dependencies via SynGCN and SemGCN, integrating these perspectives through a bi-affine fusion layer and a differential regularizer to yield a unified, disentangled representation.
    \item Dimension-Anchored Evidence Encoder: This component maps the five pedagogical dimensions into learnable semantic anchors via a prompt-based module. It employs a cross-attention mechanism to align dimension-specific rationales with the encoded text, effectively extracting evidence grounded in structured pedagogical axes.
    \item Multi-Dimensional Evaluation Head: The final module applies a shared classification head with a low-rank perturbation strategy to generate independent assessment scores for each teaching dimension. The entire pipeline is optimized through a joint supervised loss function.
\end{itemize}
The following sections elaborate on each component in detail.

\begin{table}
\centering
\caption{\textbf{TeachScope Dataset Statistics and Annotation Reliability.}
TeachScope provides a large-scale corpus of student reviews annotated with five pedagogically grounded dimensions and corresponding evidence spans.
The dataset features high inter-annotator consistency and balanced distribution across sentiment polarities.}
\label{tab:statistics}
\begin{tabular}{lccccc}
\toprule
\textbf{Category} & \textbf{Statistic} & \textbf{Train} & \textbf{Validation} & \textbf{Test} & \textbf{Total} \\
\midrule
\multicolumn{6}{l}{\textbf{Corpus Overview}} \\
\midrule
 & Number of Comments & 8,943 & 1,917 & 1,917 & 12,777 \\
 & Unique Professors & 2,690 & 1,318 & 1,293 & 5,301 \\
 & Avg. Comment Length (words) & 83.7 & 85.1 & 84.9 & 84.6 \\
 & Median Comment Length (words) & 72 & 73 & 74 & 72 \\
\midrule
\multicolumn{6}{l}{\textbf{Annotation Structure}} \\
\midrule
 & Annotated Dimensions per Comment & 5 & 5 & 5 & 5 \\
 & The maximum length of a evidence  & 281 & 264  & 280  & 281 \\
 & The minimum length of a evidence & 6 & 14 & 10 & 6 \\
 & The maximum length of a complete comment& 773 & 485 & 507 & 773 \\
 & The minimum length of a complete comment & 1 & 1 & 1 & 1 \\
\bottomrule
\end{tabular}
\end{table}

\subsection{Cross-View Graph Synergy Network}
Free-form student reviews typically manifest both syntactic regularities—such as modifier–head dependency, clause composition, and negation—and non-local semantic coherence reflecting abstract aspects of teaching quality. A single relational representation is insufficient to model both local grammatical structure and high-level semantic correlation. To address this duality, we introduce a \textit{Cross-View Graph Synergy Networ} that constructs two complementary relational graphs: a Syntactic Branch (SynGCN) to encode explicit grammatical relations via a probabilistic dependency graph, and a Semantic Branch (SemGCN) to model non-local semantic correlations through an attention-induced \cite{vaswani2017attention} affinity graph. These two graphs run in parallel to capture heterogeneous relations, and are later fused via a BiAffine information exchange to yield a structurally grounded yet semantically enriched sentence representation.

\subsubsection{Contextual Initialization with BERT}
Let a tokenized review be denoted as \(x=\{w_1,w_2,\dots,w_n\}\). Each token is passed through a frozen pretrained BERT encoder, producing contextualized token embeddings
\begin{equation}
H^{(0)} = [h^{(0)}_1,h^{(0)}_2,\dots,h^{(0)}_n] \in \mathbb{R}^{n\times d_{\text{BERT}}}, \qquad d_{\text{BERT}}=768.
\end{equation}
The encoder provides a stable linguistic prior, ensuring that the model benefits from large-scale pretraining while preventing catastrophic interference during supervised graph learning. The same initialization is shared across both branches:
\begin{equation}
H^{(0)}_{\mathrm{syn}} = H^{(0)}_{\mathrm{sem}} = H^{(0)}.
\end{equation}
This guarantees that any representational divergence in later layers arises solely from relational propagation, rather than from separate encoder updates. It also aligns token spaces, allowing later BiAffine operations to act on comparable feature manifolds.

\subsubsection{Graph Induction and Relational Propagation}
This stage constructs and propagates information through two distinct graph topologies: a syntax-based dependency graph that captures explicit grammatical structure, and a semantic affinity graph that models latent associations and topic-level dependencies within the same sentence.

\paragraph{Syntactic graph.}
For each token pair \((i,j)\), we use a probabilistic dependency parser to produce a distribution over potential arcs:
\begin{equation}
A_{\mathrm{syn}}(i,j)=\Pr((i\!\to\!j)\mid x)\in[0,1].
\end{equation}
Unlike discrete trees, this soft adjacency retains uncertainty information, mitigating parser brittleness and enhancing robustness to informal student language. The adjacency can be regularized by temperature scaling and pruning thresholding:
\begin{equation}
A_{\mathrm{syn}} \leftarrow \mathbf{1}\!\left[\frac{A_{\mathrm{syn}}}{\tau} \ge \eta \right]\odot A_{\mathrm{syn}} + I_n,
\end{equation}
where \(\tau\) controls smoothness and \(I_n\) preserves self-loops. The normalized Laplacian form
\begin{equation}
\hat{A}_{\mathrm{syn}} = D_{\mathrm{syn}}^{-1/2}A_{\mathrm{syn}}D_{\mathrm{syn}}^{-1/2},\qquad D_{\mathrm{syn}}=\mathrm{diag}(A_{\mathrm{syn}}\mathbf{1}),
\end{equation}
is then used for message passing:
\begin{equation}
H^{(\ell+1)}_{\mathrm{syn}} = \mathrm{LN}\!\left(\mathrm{GELU}\!\big(\hat{A}_{\mathrm{syn}}H^{(\ell)}_{\mathrm{syn}}W^{(\ell)}_{\mathrm{syn}}\big)\right),
\qquad W^{(\ell)}_{\mathrm{syn}}\in\mathbb{R}^{d\times d}.
\end{equation}
This propagation preserves directional dependencies while attenuating noise from uncertain arcs, allowing the model to encode grammatical constraints such as modifier scope, subordination, or conjunction relations—critical in determining sentiment targets in complex educational feedback.

\paragraph{Semantic graph.}
While syntactic propagation constrains composition according to linguistic structure, it may fail to capture long-range or thematic relations that extend beyond dependency boundaries. To supplement this, we dynamically construct a semantic adjacency matrix by scaled dot-product attention:
\begin{align}
Q &= H^{(\ell)}_{\mathrm{sem}}W_Q, \quad
K = H^{(\ell)}_{\mathrm{sem}}W_K,\\
A_{\mathrm{sem}} &= \mathrm{softmax}\!\Big(\tfrac{QK^{\top}}{\sqrt{d}}\Big)\in\mathbb{R}^{n\times n}.
\end{align}
Here, each element \(A_{\mathrm{sem}}(i,j)\) reflects the contextual correlation between tokens \(i\) and \(j\), capturing discourse-level coherence and semantic relatedness. After adding self-loops and symmetric normalization:
\begin{equation}
\tilde{A}_{\mathrm{sem}} = D^{-1/2}(A_{\mathrm{sem}} + I_n)D^{-1/2},
\end{equation}
features are updated by:
\begin{equation}
\begin{aligned}
&H^{(\ell+1)}_{\mathrm{sem}} 
= \mathrm{LN}\!\left(\mathrm{GELU}\!\big(\tilde{A}_{\mathrm{sem}}H^{(\ell)}_{\mathrm{sem}}W^{(\ell)}_{\mathrm{sem}}\big)\right), \\
&W^{(\ell)}_{\mathrm{sem}} 
\in \mathbb{R}^{d\times d}.
\end{aligned}
\end{equation}
This layer-wise adaptive graph allows information to flow along semantically coherent paths that may bypass strict syntactic boundaries, effectively capturing sentiment propagation over clause boundaries or implicit coreferences common in student evaluations.

\subsubsection{BiAffine Cross-View Fusion and Readout}
Independent processing of syntactic and semantic branches risks representational divergence, where each view overfits to its own relational prior and loses cross-modality alignment. To promote mutual refinement, we introduce a BiAffine cross-view fusion mechanism inserted after each nonlinear update. Given feature matrices \(H^{(\ell+1)}_{\mathrm{syn}}\) and \(H^{(\ell+1)}_{\mathrm{sem}}\), two asymmetric transformations are computed:
\begin{align}
H^{(\ell+1)\prime}_{\mathrm{syn}}
&=\mathrm{softmax}\!\Big(H^{(\ell+1)}_{\mathrm{syn}} W_1 (H^{(\ell+1)}_{\mathrm{sem}})^{\!\top}\Big)H^{(\ell+1)}_{\mathrm{sem}},\\
H^{(\ell+1)\prime}_{\mathrm{sem}}
&=\mathrm{softmax}\!\Big(H^{(\ell+1)}_{\mathrm{sem}} W_2 (H^{(\ell+1)}_{\mathrm{syn}})^{\!\top}\Big)H^{(\ell+1)}_{\mathrm{syn}},
\end{align}
where \(W_1, W_2 \in \mathbb{R}^{d\times d}\) are trainable weight matrices. This operation creates bidirectional alignment matrices that estimate the correspondence between syntactic and semantic token representations. Each branch receives complementary evidence from the other: syntactic nodes adjust their embeddings according to semantically inferred relevance, while semantic nodes are regularized by grammatical coherence. The updated representations are set as
\begin{equation}
H^{(\ell+1)}_{\mathrm{syn}}\leftarrow H^{(\ell+1)\prime}_{\mathrm{syn}}, \quad
H^{(\ell+1)}_{\mathrm{sem}}\leftarrow H^{(\ell+1)\prime}_{\mathrm{sem}}.
\end{equation}
Dropout is applied to prevent overfitting and encourage stable co-adaptation.

The design motivation of the BiAffine mechanism is to bridge discrete linguistic structure and continuous semantic space through soft alignment, forming an interaction field that enables information exchange while preserving each view’s inductive bias. Compared with simple concatenation or late fusion, this formulation explicitly models token-to-token correlations across relational types, yielding richer multi-view coherence and more interpretable propagation patterns.

After \(L\) stacked layers, we concatenate the final outputs of both branches to form the complete contextual representation:
\begin{equation}
H_{\mathrm{x}} = \big[\,H^{(L)}_{\mathrm{sem}}\,\|\,H^{(L)}_{\mathrm{syn}}\,\big]\in\mathbb{R}^{n\times(2d)}.
\end{equation}
This matrix retains the complementary strengths of grammatical precision and semantic coverage, serving as the foundation for the subsequent dimension-query interaction stage.

\subsection{Dimension-Anchored Evidence Encoder}

Recent studies demonstrate that introducing auxiliary information can significantly enhance a model’s representational capacity and improve task performance \cite{bei2025correlationawaregraphconvolutionalnetworks,pan2025graphnarratorgeneratingtextualexplanations}. Following this principle, we design a \emph{dimension-query mechanism} in TeachPro to integrate auxiliary semantic cues—specifically, \emph{dimension words} and \emph{prototype-guided textual snippets}—for dimension-aware supervision and interpretable reasoning. The goal is to enable each dimension query to adaptively interact with both its corresponding prototype-derived snippet group and the encoded review representation, thereby acquiring a fine-grained semantic understanding that aligns with distinct evaluation aspects. This module proceeds in three steps: (1) \emph{dimension word encoding and review-based prototype segmentation} for initialization, (2) \emph{query refinement through prototype-guided interaction}, and (3) \emph{cross-attention fusion for dimension evidence extraction}.

\subsubsection{Dimension Word Encoding and Review-Based Prototype Segmentation}
Given a tokenized review \(x=\{w_1,\dots,w_n\}\), its encoded representation from the dual-graph encoder is denoted as \(H_{\mathrm{x}}\in\mathbb{R}^{n\times(2d)}\). We define a set of \(k\) learnable dimension words
\[
W_P = [w^{(d)}_1, w^{(d)}_2, \dots, w^{(d)}_k],
\]
each representing one distinct evaluation dimension. The same frozen BERT encoder used in the previous module is applied to these dimension words to ensure consistency in the representational space:
\begin{equation}
\mathcal{Q} = \mathrm{BERT}(W_P) = [\boldsymbol{q}_1, \boldsymbol{q}_2, \dots, \boldsymbol{q}_k]^\top \in \mathbb{R}^{k\times d}.
\end{equation}

For each review \(x\), we automatically segment it into \(k\) snippets \(R=\{r_1,\dots,r_k\}\), with each snippet \(r_i\) corresponding to the textual portion most related to the \(i\)-th evaluation dimension (e.g., teaching attitude, clarity, feedback timeliness, etc.). These snippets are treated as \emph{prototype instances}—localized fragments drawn from the complete review that reflect the typical expression patterns of each dimension within the current comment. This prototype-based segmentation ensures that each dimension is grounded in authentic, review-level evidence rather than relying on manually curated rationale sets. Each snippet \(r_i=\{s_{i,1},\dots,s_{i,m_i}\}\) is then encoded through the same BERT encoder:
\begin{equation}
H_{r,i} = \mathrm{BERT}(r_i) \in \mathbb{R}^{m_i \times d}.
\end{equation}
The resulting snippet embeddings constitute fine-grained, prototype-guided evidence pools that interact with their respective dimension queries in the next stage.

\subsubsection{Prototype-Guided Interaction for Query Refinement}
In this stage, each initial query \(\boldsymbol{q}_i\) is refined through interaction with its corresponding prototype snippet representation \(H_{r,i}\). Conceptually, this process allows the query to absorb dimension-specific contextual nuances from the review itself, achieving adaptive alignment between global prior knowledge and localized prototype semantics.

Formally, we compute a relevance distribution over snippet tokens via scaled dot-product attention:
\begin{equation}
\boldsymbol{a}_i = \mathrm{softmax}\!\Big(\tfrac{H_{r,i}W_K \boldsymbol{q}_i}{\sqrt{d}}\Big) \in \mathbb{R}^{m_i},
\end{equation}
where \(W_K \in \mathbb{R}^{d\times d}\) is a trainable projection. The attention-weighted snippet representation is obtained as
\begin{equation}
\boldsymbol{p}_i = \boldsymbol{a}_i^\top H_{r,i} \in \mathbb{R}^{d},
\end{equation}
capturing the most relevant prototype semantics for the \(i\)-th dimension. To fuse this information with the prior dimension word embedding, we apply a gated update:
\begin{equation}
\lambda_i = \sigma\!\big(\boldsymbol{u}^\top [\boldsymbol{q}_i; \boldsymbol{p}_i]\big), \qquad
\boldsymbol{q}^\star_i = (1-\lambda_i)\boldsymbol{q}_i + \lambda_i \boldsymbol{p}_i,
\end{equation}
where \(\boldsymbol{u} \in \mathbb{R}^{2d}\) is learnable and \(\sigma\) denotes the sigmoid function. This mechanism ensures that each query incorporates prototype-grounded semantic evidence while retaining its original dimensional intent. Collecting across all dimensions yields
\begin{equation}
\mathcal{Q}_{\mathrm{new}} = [\boldsymbol{q}^\star_1, \boldsymbol{q}^\star_2, \dots, \boldsymbol{q}^\star_k]^\top \in \mathbb{R}^{k\times d}.
\end{equation}
Intuitively, this stage performs dynamic grounding of each dimension word through prototype evidence extracted from the review, enabling fine-grained differentiation across evaluation aspects.

\subsubsection{Cross-Attention Fusion for Dimension Evidence Extraction}
After refinement, the updated query matrix \(\mathcal{Q}_{\mathrm{new}}\) interacts with the encoded full-review representation \(H_{\mathrm{x}}\) to extract dimension-specific evidence vectors. Before performing cross-attention, we stabilize the refined queries using the DyT normalization function \cite{zhu2025transformersnormalization}, which compresses outliers and accelerates convergence:
\begin{equation}
H_Q = \mathrm{DyT}(\mathcal{Q}_{\mathrm{new}}), \qquad 
\mathrm{DyT}(\boldsymbol{z}) = \boldsymbol{\gamma} \odot \tanh(\alpha \boldsymbol{z}) + \boldsymbol{\beta},
\end{equation}
where \(\boldsymbol{\gamma}, \boldsymbol{\beta} \in \mathbb{R}^d\) and \(\alpha>0\) are learnable parameters, and \(\odot\) denotes element-wise multiplication.

We then compute query-conditioned attention over all review tokens:
\begin{equation}
A = \mathrm{softmax}\!\Big(\tfrac{H_Q H_{\mathrm{x}}^\top}{\sqrt{d}}\Big) \in \mathbb{R}^{k\times n},
\qquad
H_E = A H_{\mathrm{x}} \in \mathbb{R}^{k\times d}.
\end{equation}
Here, the \(i\)-th row of \(H_E\) corresponds to the attention-aggregated evidence vector for the \(i\)-th evaluation dimension. Each vector encapsulates the most discriminative textual evidence that supports the assessment of that dimension within the review. This cross-attention operation effectively performs query-conditioned reasoning, aligning textual semantics with dimension-specific expectations and producing structured embeddings for the downstream classification module. The resulting \(H_E\) serves as the final dimension-aware representation of the review and constitutes the foundation for generating interpretable and fine-grained evaluation predictions in the subsequent module.

\subsection{Parameter-Efficient Multi-Dimensional Evaluation Prediction Head }

In multidimensional evaluation tasks, each dimension represents an independent semantic perspective, which ideally requires its own classifier. However, training multiple classifiers separately not only leads to substantial parameter redundancy but also weakens the model’s ability to maintain a coherent global feature space. To address this, we design a unified prediction architecture that retains shared parameters across dimensions while introducing lightweight, dimension-adaptive perturbations. This structure ensures that all evaluation dimensions share a consistent projection foundation while still capturing fine-grained, dimension-specific discriminative cues.

\paragraph{Unified Low-Rank Perturbation for Classification.}
Given the dimension-aware representations \(H_E \in \mathbb{R}^{k \times d}\) from the previous module, where each row \(\boldsymbol{h}_i \in \mathbb{R}^{d}\) denotes the evidence vector for the \(i\)-th evaluation dimension, the prediction head introduces per-dimension scaling and bias factors to enable flexible yet compact specialization. A shared projection matrix \(W_P \in \mathbb{R}^{d \times d}\) is maintained for all dimensions to ensure a global transformation pattern, while each dimension is equipped with small, learnable perturbation parameters \(\boldsymbol{z}_i, \boldsymbol{s}_i, \boldsymbol{b}_i \in \mathbb{R}^{d}\). These parameters are collected into the matrices \(Z, S, B \in \mathbb{R}^{k \times d}\). 

The overall transformation for the \(i\)-th dimension can be expressed as
\begin{equation}
\label{eq:perturb}
\mathrm{Project}_i(\boldsymbol{h}_i) = \Big((\boldsymbol{h}_i \odot \boldsymbol{z}_i)W_P\Big) \odot \boldsymbol{s}_i + \boldsymbol{b}_i,
\end{equation}
where \(\odot\) denotes element-wise multiplication. This operation first scales the input representation \(\boldsymbol{h}_i\) according to a dimension-specific factor \(\boldsymbol{z}_i\), applies a shared linear projection, and then adjusts the resulting feature through a second scaling \(\boldsymbol{s}_i\) and a bias term \(\boldsymbol{b}_i\). This low-rank perturbation allows each dimension to learn fine-tuned discriminative mappings while reusing the global feature backbone, thereby avoiding the need for multiple independent classifiers \cite{gorishniy2025tabmadvancingtabulardeep}.

Each transformed representation \(\mathrm{Project}_i(\boldsymbol{h}_i)\) is passed through a nonlinearity and a dropout layer to enhance robustness:
\begin{equation}
\boldsymbol{u}_i = \mathrm{Dropout}\!\big(\mathrm{GELU}(\mathrm{Project}_i(\boldsymbol{h}_i))\big),
\end{equation}
and then projected to the output logits space by a shared classifier \(W_C \in \mathbb{R}^{d \times m}\) with bias \(\boldsymbol{c} \in \mathbb{R}^{m}\):
\begin{equation}
\boldsymbol{o}_i = \boldsymbol{u}_i W_C + \boldsymbol{c}, \qquad
\hat{\boldsymbol{y}}_i = \mathrm{softmax}(\boldsymbol{o}_i) \in \mathbb{R}^{m},
\end{equation}
where \(m\) denotes the number of polarity categories per dimension. This design ensures that each evaluation dimension maintains independent decision flexibility while being constrained by a shared projection geometry, striking a balance between specialization and generalization.

The TeachPro model is optimized using a cross-entropy loss averaged over all evaluation dimensions. Given the ground-truth polarity labels \(\boldsymbol{y}_i \in \{0,1\}^m\), the overall objective function is defined as
\begin{equation}
\label{eq:loss}
\mathcal{L} = \frac{1}{k} \sum_{i=1}^{k} \Big(- \sum_{c=1}^{m} \omega_c\, y_{i,c}\, \log \hat{y}_{i,c}\Big),
\end{equation}
where \(\omega_c > 0\) denotes optional class weights that mitigate label imbalance across sentiment categories.  
This loss jointly optimizes both shared and dimension-specific parameters, enabling the model to learn coherent, fine-grained representations that accurately capture multidimensional evaluation semantics.

\section{Experiments}
This section presents a comprehensive empirical study designed to evaluate the behaviour, robustness, and interpretability of models on the proposed multi-dimensional instructional assessment task. The experiments are organized to progressively examine: (i) how a diverse collection of foundational neural architectures perform under a unified and controlled setting; (ii) how the components of the proposed framework contribute to its overall effectiveness through systematic removal of key modules; (iii) how the low-rank perturbation mechanism influences representational structure and prediction behaviour; (iv) how dimension-specific embeddings evolve across network layers and encode inter-facet relationships; and (v) whether the model's facet predictions are grounded in textual evidence through alignment between learned attention patterns and annotated rationale spans. Together, these analyses provide a multi-perspective evaluation that covers predictive accuracy, architectural contributions, representational dynamics, and evidence-based reasoning.

In what follows, we first detail the implementation settings adopted for both the proposed model and all baseline systems, and subsequently present the experimental results in the order outlined above.

\subsection{Scope of the Comparison and Implementation Details}
Our study emphasizes relative architectural behaviour rather than exhaustive per-model tuning. To obtain a fair and reproducible comparison, all baselines and the proposed TeachPro model are trained under a unified experimental protocol. The dataset is split into training, validation, and test partitions with a fixed ratio of 70\%, 15\%, and 15\%, respectively, using a shared random seed. Input comments are tokenized to a maximum length of 128, and all models operate on the same token sequences so that they learn from an identical feature space.

Training is performed with a batch size of 64 and up to 10 epochs for every model. Unless otherwise specified, all networks are optimized using the Adam optimizer with an initial learning rate of $2\times10^{-3}$ and a standard cross-entropy loss for the three-way polarity classification on each of the five dimensions. A cosine-annealing learning-rate schedule is applied consistently across models, with $T_{\max}=10$ and a minimum learning rate of $5\times10^{-4}$, ensuring that training dynamics are governed by the same decay pattern. The proposed TeachPro model uses a hidden dimensionality of 768 and a dropout rate of 0.1, matching the capacity of the Transformer-based encoders it is compared with. All experiments are conducted with a fixed random seed for data splitting and initialization, and each configuration is repeated five times with different seeds. Reported results correspond to the mean performance across runs, ensuring robustness against stochastic variation.

\subsection{Baseline Comparison with Deep Neural Models}
\begin{table}
\centering
\caption{Validation performance of baseline models on the last training epoch.}
\label{tab:baseline_val}
\setlength{\tabcolsep}{3pt}
\renewcommand{\arraystretch}{1.1}
\begin{tabular*}{\linewidth}{@{\extracolsep{\fill}}l*{6}{cc}@{}}
\toprule
\multirow{2}{*}{Model} &
\multicolumn{2}{c}{Professionalism} &
\multicolumn{2}{c}{Occupational} &
\multicolumn{2}{c}{Effectiveness} &
\multicolumn{2}{c}{Quality} &
\multicolumn{2}{c}{Other} &
\multicolumn{2}{c}{Average} \\
\cmidrule(lr){2-3}
\cmidrule(lr){4-5}
\cmidrule(lr){6-7}
\cmidrule(lr){8-9}
\cmidrule(lr){10-11}
\cmidrule(lr){12-13}
& Acc & F1 & Acc & F1 & Acc & F1 & Acc & F1 & Acc & F1 & Acc & F1 \\
\midrule
LSTM       & 0.4737 & 0.2143 & 0.5164 & 0.2271 & 0.4731 & 0.2153 & 0.6813 & 0.2701 & 0.5968 & 0.2492 & 0.5483 & 0.2352 \\
TextCNN    & 0.7240 & 0.7217 & 0.7199 & 0.6870 & 0.6604 & 0.6536 & 0.7903 & 0.6392 & 0.7329 & 0.6763 & 0.7255 & 0.6755 \\
BERT       & 0.4737 & 0.2143 & 0.5170 & 0.2272 & 0.4726 & 0.2140 & 0.6813 & 0.2701 & 0.5968 & 0.2492 & 0.5483 & 0.2349 \\
DistilBERT & 0.4737 & 0.2143 & 0.5170 & 0.2272 & 0.4726 & 0.2140 & 0.6813 & 0.2701 & 0.5968 & 0.2492 & 0.5483 & 0.2349 \\
RoBERTa    & 0.4737 & 0.2143 & 0.5170 & 0.2272 & 0.4726 & 0.2140 & 0.6813 & 0.2701 & 0.5968 & 0.2492 & 0.5483 & 0.2349 \\
SpanBERT   & 0.4737 & 0.2143 & 0.5170 & 0.2272 & 0.4726 & 0.2140 & 0.6813 & 0.2701 & 0.5968 & 0.2492 & 0.5483 & 0.2349 \\
StructBERT & 0.4737 & 0.2143 & 0.5170 & 0.2272 & 0.4726 & 0.2140 & 0.6813 & 0.2701 & 0.5968 & 0.2492 & 0.5483 & 0.2349 \\
ELECTRA    & 0.4737 & 0.2143 & 0.5170 & 0.2272 & 0.4726 & 0.2140 & 0.6813 & 0.2701 & 0.5968 & 0.2492 & 0.5483 & 0.2349 \\
BART       & 0.4737 & 0.2143 & 0.5170 & 0.2272 & 0.4726 & 0.2140 & 0.6813 & 0.2701 & 0.5968 & 0.2492 & 0.5483 & 0.2349 \\
TinyBERT   & 0.6312 & 0.6455 & 0.7162 & 0.6805 & 0.6030 & 0.5182 & 0.7903 & 0.5488 & 0.7360 & 0.6593 & 0.6954 & 0.6105 \\
\midrule
\textbf{Ours}       & \textbf{0.8122} & \textbf{0.8083} & \textbf{0.7955} & \textbf{0.7768} & \textbf{0.7673} & \textbf{0.7540} & \textbf{0.8555} & \textbf{0.7556} & \textbf{0.7939} & \textbf{0.7411} & \textbf{0.8049} & \textbf{0.7672} \\
\midrule
           & QWK    & ECE    & QWK    & ECE    & QWK    & ECE    & QWK    & ECE    & QWK    & ECE    & QWK    & ECE    \\
\midrule
LSTM       & 0.0003 & 0.0066 & 0.0005 & 0.0291 & 0.0012 & 0.0278 & 0.0000 & 0.0140 & 0.0000 & 0.0151 & 0.0003 & 0.0185 \\
TextCNN    & 0.6916 & 0.1598 & 0.7380 & 0.1649 & 0.6687 & 0.1588 & 0.7196 & 0.1136 & 0.6017 & 0.1256 & 0.6839 & 0.1445 \\
BERT       & 0.0000 & 0.0058 & 0.0000 & 0.0264 & 0.0000 & 0.0344 & 0.0000 & 0.0150 & 0.0000 & 0.0348 & 0.0000 & 0.0233 \\
DistilBERT & 0.0000 & 0.0035 & 0.0000 & 0.0022 & 0.0000 & 0.0323 & 0.0000 & 0.0070 & 0.0000 & 0.0176 & 0.0000 & 0.0125 \\
RoBERTa    & 0.0000 & 0.0181 & 0.0000 & 0.0100 & 0.0000 & 0.0418 & 0.0000 & 0.0148 & 0.0000 & 0.0356 & 0.0000 & 0.0241 \\
SpanBERT   & 0.0000 & 0.0020 & 0.0000 & 0.0062 & 0.0000 & 0.0239 & 0.0000 & 0.0254 & 0.0000 & 0.0270 & 0.0000 & 0.0169 \\
StructBERT & 0.0000 & 0.0221 & 0.0000 & 0.0301 & 0.0000 & 0.0348 & 0.0000 & 0.0043 & 0.0000 & 0.0009 & 0.0000 & 0.0184 \\
ELECTRA    & 0.0000 & 0.0047 & 0.0000 & 0.0402 & 0.0000 & 0.0011 & 0.0000 & 0.0209 & 0.0000 & 0.0215 & 0.0000 & 0.0177 \\
BART       & 0.0000 & 0.0128 & 0.0000 & 0.0097 & 0.0000 & 0.0325 & 0.0000 & 0.0054 & 0.0000 & 0.0198 & 0.0000 & 0.0160 \\
TinyBERT   & 0.5788 & 0.1289 & 0.7333 & 0.1911 & 0.5481 & 0.0698 & 0.6525 & 0.0368 & 0.5766 & 0.1212 & 0.6179 & 0.1096 \\
\midrule
\textbf{Ours}       & \textbf{0.8117} & \textbf{0.0634} & \textbf{0.8278} & \textbf{0.0483} & \textbf{0.7758} & \textbf{0.0591} & \textbf{0.8479} & \textbf{0.0320} & \textbf{0.6772} & \textbf{0.0408} & \textbf{0.7881} & \textbf{0.0487} \\
\bottomrule
\end{tabular*}
\end{table}

To comprehensively evaluate the proposed model, we compare it against a set of representative deep neural baselines, including a bidirectional LSTM and a convolutional architecture (TextCNN), as well as a range of pre-trained Transformer encoders: BERT~\cite{DBLP:journals/corr/abs-1810-04805}, DistilBERT~\cite{sanh2020distilbertdistilledversionbert}, RoBERTa~\cite{liu2019robertarobustlyoptimizedbert}, SpanBERT~\cite{joshi2020spanbertimprovingpretrainingrepresenting}, StructBERT~\cite{wang2019structbertincorporatinglanguagestructures}, ELECTRA~\cite{clark2020electrapretrainingtextencoders}, BART~\cite{lewis2019bartdenoisingsequencetosequencepretraining}, and TinyBERT~\cite{jiao2020tinybertdistillingbertnatural}. These baselines cover both recurrent and convolutional sequence models as well as modern pre-trained language models with diverse pre-training objectives. The validation results on the last training epoch are summarized in Table~\ref{tab:baseline_val}, where we report accuracy and F1 for each instructional facet, together with the Quadratic Weighted Kappa (QWK) and Expected Calibration Error (ECE) to jointly assess discriminative performance and calibration quality.

As shown in the upper block of Table~\ref{tab:baseline_val}, the proposed model substantially outperforms all baselines across all facets and on the averaged metrics. In terms of average accuracy and F1, our method achieves 0.8049 and 0.7672, respectively, whereas the strongest baselines (TextCNN and TinyBERT) remain in the range of approximately 0.70--0.73 accuracy and 0.61--0.68 F1. The gap is consistent across dimensions: for each instructional facet, the proposed approach yields higher accuracy and F1 than any of the compared models. This indicates that simply fine-tuning generic sentence encoders, even strong pre-trained Transformers, is insufficient to capture the fine-grained, facet-specific distinctions required by the multi-dimensional assessment task.

The lower block of Table~\ref{tab:baseline_val} further highlights the advantage of the proposed model when considering QWK and ECE. Our method obtains the highest average QWK, reflecting superior agreement with the ground-truth facet ratings, while maintaining a moderate ECE that indicates reasonably calibrated probability estimates. In contrast, several Transformer-based baselines exhibit extremely low QWK values that are close to random agreement, despite sometimes reporting small ECE values. This combination suggests that their predictions are poorly discriminative and that the apparent calibration is largely a consequence of producing near-uniform or low-variance scores rather than meaningful alignment with the label distributions. Models such as TextCNN and TinyBERT achieve stronger QWK than other baselines but still fall noticeably short of the proposed approach, and their calibration remains substantially worse than that of our model.

An important observation from these results is that most baseline models struggle to effectively exploit shared textual representations for multiple instructional dimensions. In the baseline setting, a single encoder must support several independent classifiers without explicit access to dimension-specific semantic priors or structured guidance on how different facets should be disentangled. This configuration makes multi-facet prediction prone to interference between dimensions and limits the model’s ability to associate particular segments of the comment with the corresponding instructional criteria. By contrast, our approach introduces explicit dimension embeddings and query refinement mechanisms that act as facet-aware prompts, enabling the model to construct distinct yet related representations for each instructional dimension and to attend to facet-relevant evidence in the input text. The consistently higher accuracy, F1, and QWK, together with improved calibration, provide quantitative support for the effectiveness of this design in addressing the challenges of multi-dimensional instructional assessment.

\subsection{Ablation Study}

\begin{table}
\centering
\caption{Ablation study on DualGCN and Query Refinement modules.}
\label{tab:ablation_dualgcn_queryrefine}
\setlength{\tabcolsep}{3pt}
\renewcommand{\arraystretch}{1.1}
\begin{tabular*}{\linewidth}{@{\extracolsep{\fill}}l*{6}{cc}@{}}
\toprule
\multirow{2}{*}{Method} &
\multicolumn{2}{c}{Professionalism} &
\multicolumn{2}{c}{Occupational} &
\multicolumn{2}{c}{Effectiveness} &
\multicolumn{2}{c}{Quality} &
\multicolumn{2}{c}{Other} &
\multicolumn{2}{c}{Average} \\
\cmidrule(lr){2-3}
\cmidrule(lr){4-5}
\cmidrule(lr){6-7}
\cmidrule(lr){8-9}
\cmidrule(lr){10-11}
\cmidrule(lr){12-13}
& Acc & F1 & Acc & F1 & Acc & F1 & Acc & F1 & Acc & F1 & Acc & F1 \\
\midrule
w/o DualGCN          & 0.7955 & 0.7936 & 0.7986 & 0.7633 & 0.7512 & 0.7374 & 0.8644 & 0.7523 & 0.7908 & 0.7479 & 0.8001 & 0.7589 \\
w/o Query Refinement & 0.7981 & 0.8003 & 0.7705 & 0.7181 & 0.6813 & 0.6298 & 0.8425 & 0.6603 & 0.7799 & 0.7412 & 0.7744 & 0.7100 \\
Ours                 & 0.8122 & 0.8083 & 0.7955 & 0.7768 & 0.7673 & 0.7540 & 0.8555 & 0.7556 & 0.7939 & 0.7411 & 0.8049 & 0.7672 \\
\midrule
                     & QWK    & ECE    & QWK    & ECE    & QWK    & ECE    & QWK    & ECE    & QWK    & ECE    & QWK    & ECE    \\
\midrule
w/o DualGCN          & 0.8029 & 0.0715 & 0.8362 & 0.0749 & 0.7746 & 0.0474 & 0.8454 & 0.0378 & 0.6898 & 0.0154 & 0.7898 & 0.0494 \\
w/o Query Refinement & 0.7938 & 0.0189 & 0.8099 & 0.0625 & 0.7040 & 0.0193 & 0.7910 & 0.0435 & 0.6897 & 0.0370 & 0.7577 & 0.0362 \\
Ours                 & 0.8117 & 0.0634 & 0.8278 & 0.0483 & 0.7758 & 0.0591 & 0.8479 & 0.0320 & 0.6772 & 0.0408 & 0.7881 & 0.0487 \\
\bottomrule
\end{tabular*}
\end{table}

To evaluate the contribution of the main architectural components, we conduct an ablation study in which two critical modules are selectively removed from the full framework. This design allows us to isolate the functional role of each component and to examine whether the proposed mechanisms are necessary for achieving reliable and fine-grained instructional facet assessment. Specifically, the configuration denoted as \emph{w/o DualGCN} removes the cross-view graph collaborative network, thereby preventing the model from integrating syntactic dependencies and semantic associations when propagating token representations. In this setting, the encoder relies solely on the contextualized embeddings produced by the frozen language model, without access to the structured relational information that the dual graph formulation is designed to capture. The configuration denoted as \emph{w/o Query Refinement} removes the prototype-guided dynamic update of the facet queries; consequently, the five instructional dimension embeddings remain static throughout the process and are not adapted to comment-specific linguistic patterns. This ablation examines whether dynamically aligning facet queries with prototype segments from each comment is essential for capturing nuanced lexical cues associated with different teaching aspects. Experimental results are summarized in Table~\ref{tab:ablation_dualgcn_queryrefine}.

Across all facets, the full model achieves the strongest overall performance. Compared with \emph{w/o DualGCN}, the complete system obtains absolute improvements of approximately 1.5\% in average accuracy and 1.1\% in average F1. The gains are consistent across facets, with notable enhancements on Effectiveness, where accuracy increases from 75.12\% to 76.73\% (+1.61\%) and F1 increases from 73.74\% to 75.40\% (+1.66\%). These results indicate that the dual-graph propagation mechanism contributes meaningfully to the model's ability to capture how syntactic structure and semantic coherence jointly influence students' descriptions of teaching performance. In the absence of this module, the encoder lacks structural guidance, leading to attenuated discrimination when comments contain implicit judgments, long-span contextual cues, or complex sentiment compositions.

The effect of removing Query Refinement is even more pronounced. Relative to the full model, \emph{w/o Query Refinement} results in a decrease of roughly 3.0\% in average accuracy and 5.7\% in average F1. The most substantial degradation again appears in the Effectiveness facet, where accuracy falls from 76.73\% to 68.13\% (–8.60\%) and F1 drops from 75.40\% to 62.98\% (–12.42\%). Considerable reductions also occur on Quality and the overall average score. These findings suggest that static facet embeddings are insufficient for capturing the diverse lexical expressions used by students to evaluate specific teaching dimensions. The prototype-guided refinement mechanism enables the facet queries to adjust dynamically to the latent semantic patterns present in each comment, allowing the model to better differentiate between subtle positive, neutral, and negative judgments associated with each instructional aspect. Without this dynamic alignment, the model tends to rely on overly generic query representations, leading to reduced sensitivity to facet-specific discourse signals. 
The QWK and ECE metrics provide additional evidence. The full model attains the highest average QWK (78.81\%), outperforming \emph{w/o DualGCN} by approximately 1.82\% and \emph{w/o Query Refinement} by 3.52\%. This indicates that both modules jointly improve the model's ability to produce ordinally consistent predictions that align well with human annotations. Calibration, as measured by ECE, remains stable across variants. Although \emph{w/o Query Refinement} exhibits slightly lower ECE on some facets, these differences are relatively minor and do not compensate for the marked decreases in accuracy, F1, and QWK. This pattern suggests that the main effect of Query Refinement lies in strengthening discriminative power and ordinal consistency rather than altering confidence distributions.

Overall, the ablation results demonstrate that the two examined components are complementary. DualGCN enhances relational representation learning by integrating structural information that is not directly available from the language model, while Query Refinement ensures that facet-level queries adapt to the contextual characteristics of each comment. Their joint contribution is necessary to achieve the full model's capability in fine-grained multi-facet instructional evaluation.

\subsection{Evaluation of the Low-Rank Perturbation Mechanism}

\begin{table}
\centering
\caption{Comparison of resource consumption and efficiency between the independent classification head and the low-rank shared head during training and inference (batch size = 16).}
\label{tab:lowrank_evaluation}
\setlength{\tabcolsep}{3pt}
\renewcommand{\arraystretch}{1.1}
\begin{tabular*}{\linewidth}{@{\extracolsep{\fill}}lcccc@{}}
\toprule
\multicolumn{5}{c}{\textbf{Training Stage: Resource and Efficiency}} \\
\midrule
\textbf{Method} & \textbf{Parameters} & \textbf{Time (ms)} & \textbf{Samples/sec} & \textbf{Peak GPU Mem} \\
\midrule
Independent   & 4{,}602{,}895 & 10.5781 & 94.53   & 102.84 MB \\
Shared        & 943{,}363    & 6.0928  & 164.13  & 74.49 MB  \\
\midrule
\multicolumn{5}{c}{\textbf{Inference Stage: Resource and Efficiency}} \\
\midrule
\textbf{Method} & \textbf{Parameters} & \textbf{Time (ms)} & \textbf{Samples/sec} & \textbf{Peak GPU Mem} \\
\midrule
Independent   & 4{,}602{,}895 & 1.5085 & 662.93  & 65.28 MB \\
Shared       & 943{,}363    & 0.8199 & 1219.59 & 65.37 MB \\
\bottomrule
\end{tabular*}
\end{table}

To evaluate the efficiency of the proposed low-rank perturbation mechanism, we compare the shared low-rank classification head with a variant that uses independent classification heads for each instructional facet. In the independent configuration, denoted as \emph{Independent}, five separate classification heads are trained, one per facet, without any parameter sharing across facets. In the proposed configuration, denoted as \emph{Shared}, a single shared projection with low-rank facet-specific perturbations is employed, so that most parameters are shared while only a small set of perturbation parameters is specific to each facet. The backbone encoder and all other components are kept identical, and the batch size is fixed to 16 in all runs to ensure a fair comparison. The resource consumption and efficiency statistics during training and inference are reported in Table~\ref{tab:lowrank_evaluation}.

During training, the independent configuration requires 4{,}602{,}895 parameters, whereas the shared low-rank configuration uses 943{,}363 parameters, corresponding to a reduction of approximately 79.5\% in the number of parameters attributed to the classification layer. This parameter reduction is reflected in both computational cost and memory usage. The average time per training iteration decreases from 10.5781\,ms for the independent heads to 6.0928\,ms for the shared low-rank head, which is a reduction of about 42.4\%. Consistently, the throughput increases from 94.53 to 164.13 samples per second, representing an improvement of approximately 73.6\%. Peak GPU memory consumption is reduced from 102.84\,MB to 74.49\,MB, a decrease of about 27.6\%. These results indicate that replacing fully independent heads with the low-rank shared design substantially improves training efficiency while reducing the memory footprint.

A similar pattern is observed at inference time. The independent configuration achieves an average forward time of 1.5085\,ms per batch, whereas the shared low-rank configuration requires 0.8199\,ms, which corresponds to a reduction of approximately 45.6\% in inference latency. Throughput increases from 662.93 to 1219.59 samples per second, an improvement of about 84.0\%. The peak GPU memory consumption during inference remains effectively unchanged, with 65.28\,MB for the independent heads and 65.37\,MB for the shared low-rank head, indicating that the low-rank perturbation mechanism introduces negligible additional memory overhead at inference time. Overall, these measurements indicate that the low-rank shared classification head offers a substantially more efficient parameterization than the use of fully independent heads. The reduction in parameter count directly translates into lower computational cost, faster training and inference, and reduced memory usage during training, all while preserving the model’s ability to operate effectively across multiple instructional facets. The empirical results therefore support the use of low-rank perturbations as a resource-efficient design choice for multi-facet classification architectures.

\begin{figure*}
    \centering
    \includegraphics[width=1.0\textwidth]{ 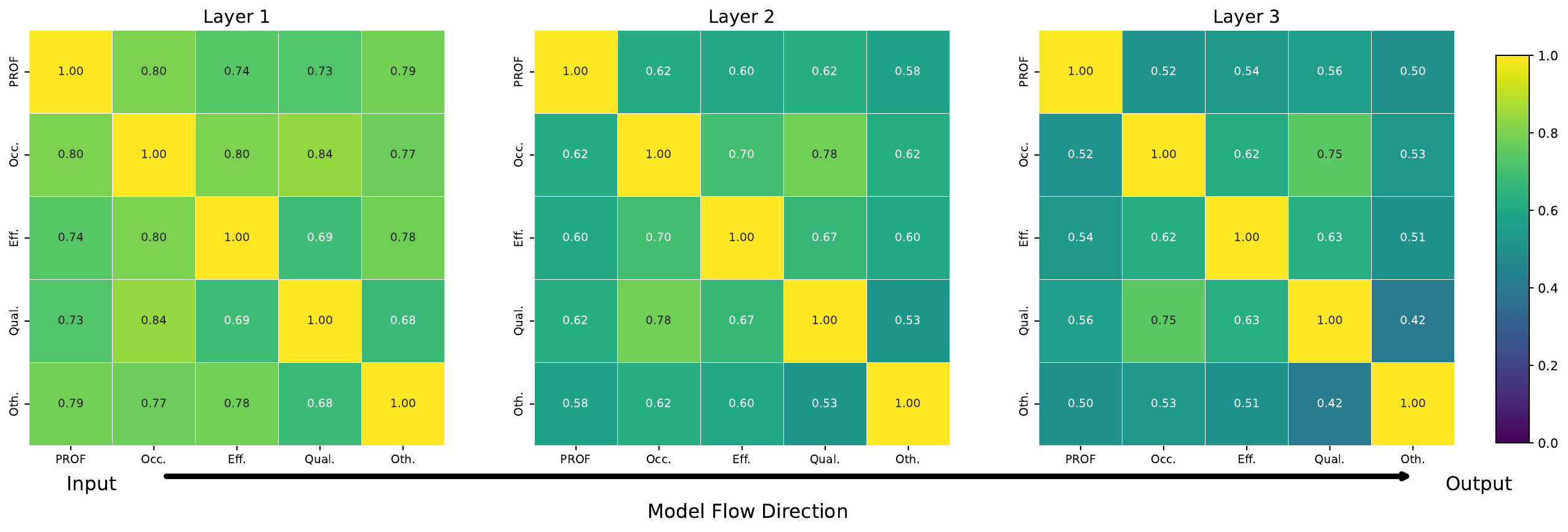}
    \caption{How inter-dimensional correlation changes as model depth increases}
    \label{fig:3pdf}
\end{figure*}

\subsection{The Evolution of Dimension Embedding Similarity Across Network Layers}

To examine how statistical dependencies among instructional dimensions are processed within the network, we relate empirical co-occurrence frequencies from the dataset to the pairwise similarities among the learned dimension embeddings across successive layers, as visualized in Figure~\ref{fig:3pdf}. Each heatmap reports the cosine similarity between the embeddings associated with the five instructional facets, with rows corresponding to input dimensions and columns to output dimensions at a given layer. This perspective enables us to trace how initially entangled facet representations are progressively reorganized as the model depth increases.

In the first layer, the similarity matrix is dominated by uniformly high off-diagonal values, with most inter-dimensional similarities concentrated near the upper end of the scale. This pattern indicates that all facets begin in a highly entangled representational space in which the model has not yet formed clear dimension-specific distinctions. At this stage, the latent dimensional space primarily reflects coarse aggregations of global co-occurrence tendencies, and the representations provide limited discriminative structure for distinguishing facets that frequently co-occur from those that should remain independent in the prediction process.

In the second layer, the correlation structure becomes more differentiated. Off-diagonal similarities systematically decrease, and the matrix exhibits greater variability across facet pairs. Some relationships remain relatively strong, whereas others are substantially weakened. This progression suggests that the network begins to selectively preserve dependencies supported by the empirical data distribution while attenuating spurious couplings introduced in earlier cross-attention interactions. The resulting topology therefore corresponds to a more structured representation in which only a subset of inter-dimensional relationships remains tightly coupled.

By the third layer, inter-dimensional similarities are further reduced and the matrix becomes more diagonally dominant. Most off-diagonal entries fall within a moderate range, while only a small number of robust associations persist. This indicates that the final mapping stage functions as a refinement mechanism that stabilizes the dimensional space, promoting approximate conditional independence among most facets while retaining the strongest and most semantically grounded dependencies consistent with observed co-occurrence statistics. Overall, the progression from Layer~1 to Layer~3 demonstrates that the model does not simply reproduce raw co-occurrence patterns; rather, it progressively factorizes the facet embedding space, disentangling dimensional representations while selectively preserving the most informative inter-facet relationships.

\subsection{Evidence Alignment between Dimension Attention and Annotated Rationale Spans}
\begin{figure*}
    \centering

    \includegraphics[width=0.6\linewidth]{ 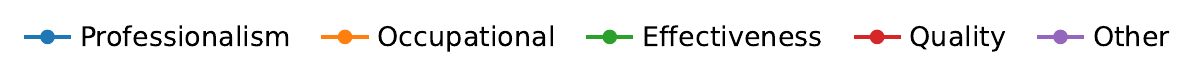}

    \begin{subfigure}{0.245\linewidth}
        \centering
        \includegraphics[width=\linewidth]{ 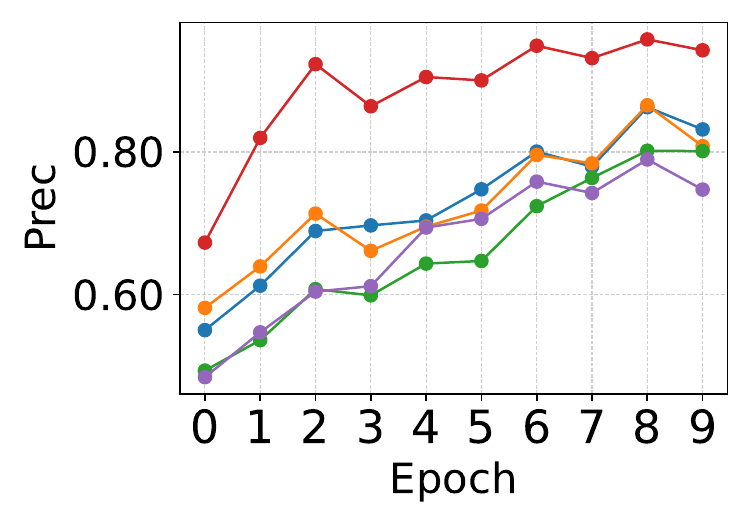}
        \caption{Precision vs Epoch}
    \end{subfigure}
    \hfill
    \begin{subfigure}{0.245\linewidth}
        \centering
        \includegraphics[width=\linewidth]{ 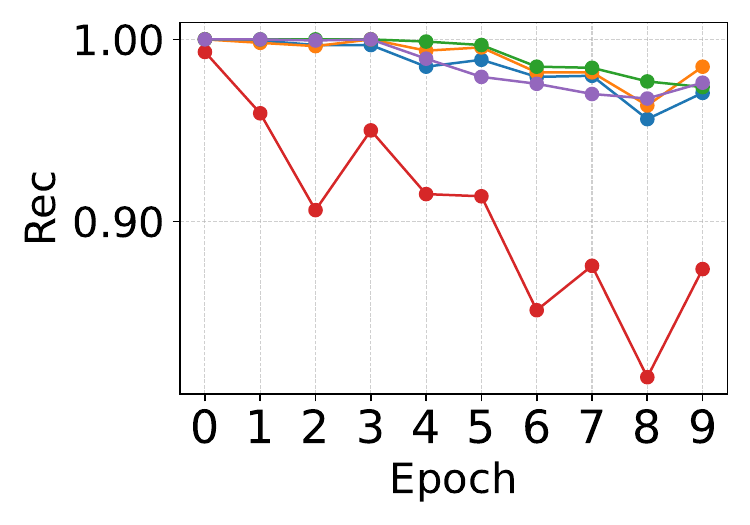}
        \caption{Recall vs Epoch}
    \end{subfigure}
    \hfill
    \begin{subfigure}{0.245\linewidth}
        \centering
        \includegraphics[width=\linewidth]{ 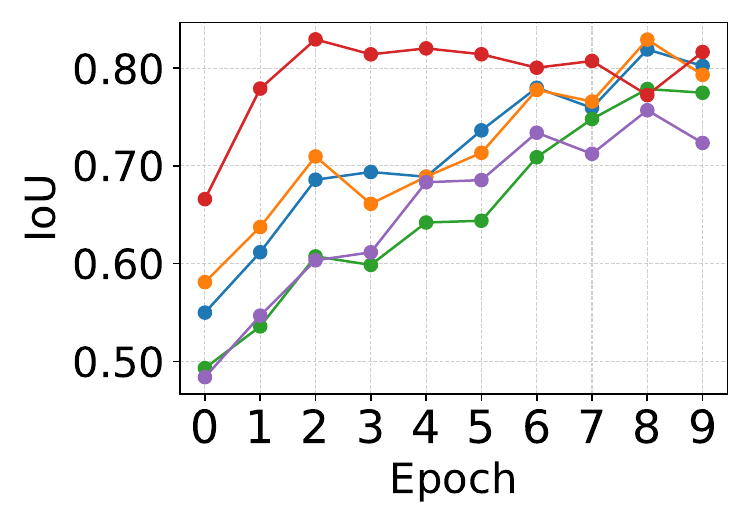}
        \caption{IoU vs Epoch}
    \end{subfigure}
    \hfill
    \begin{subfigure}{0.245\linewidth}
        \centering
        \includegraphics[width=\linewidth]{ 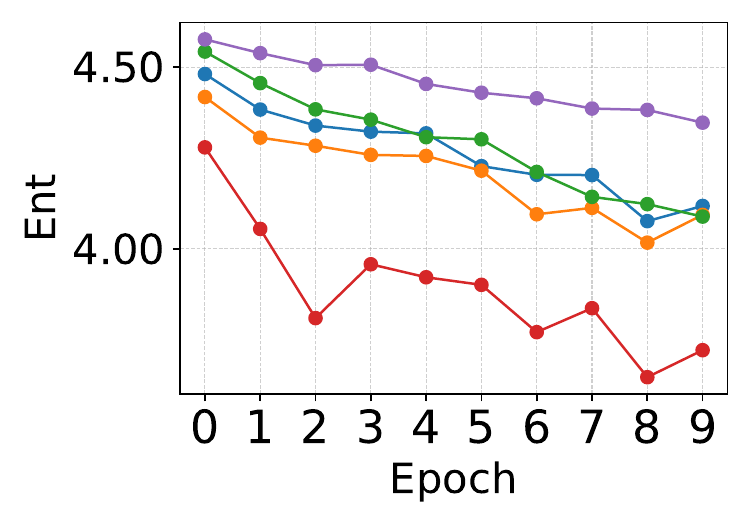}
        \caption{Entropy vs Epoch}
    \end{subfigure}

    \caption{
        Evidence alignment analysis across training epochs. 
        (a) Precision measures the proportion of model-highlighted tokens that correspond to annotated evidence; 
        (b) Recall measures the proportion of annotated evidence covered by model attention; 
        (c) IoU quantifies the overlap between predicted and gold evidence spans; 
        and (d) entropy reflects the concentration of attention, with lower entropy indicating more focused evidence attribution.
    }
    \label{fig:evidence_alignment}
\end{figure*}
\begin{figure*}
    \centering
    \begin{subfigure}{0.95\linewidth}
        \centering
        \includegraphics[width=\linewidth]{ 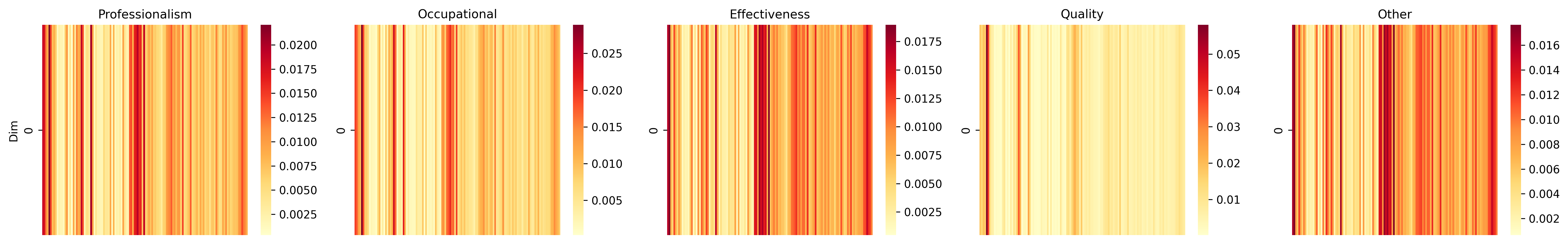}
        \caption{Attention distribution of dimension embeddings over tokens at an early training epoch.}
    \end{subfigure}
    
    \vspace{0.6em}
    
    \begin{subfigure}{0.95\linewidth}
        \centering
        \includegraphics[width=\linewidth]{ 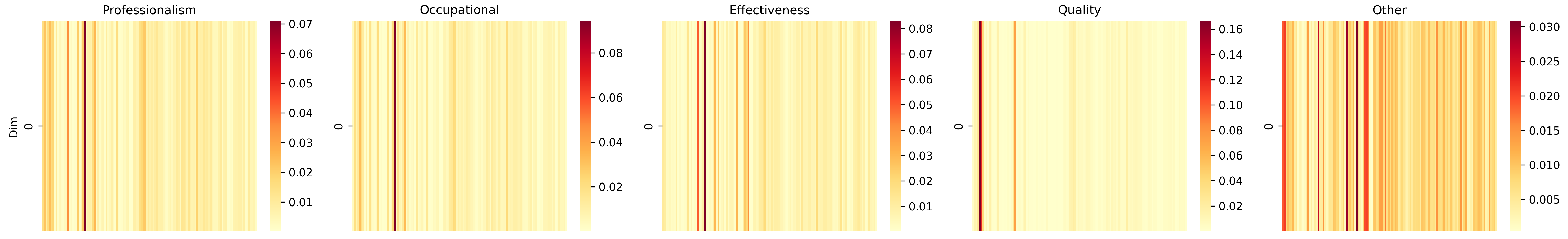}
        \caption{Attention distribution of dimension embeddings over tokens at the final training epoch.}
    \end{subfigure}
    \caption{Visualization of token-level attention weights from each instructional dimension embedding to the input comment. Each panel shows, for a fixed epoch, how attention mass is allocated over tokens for the five dimensions, with warmer colors indicating higher normalized attention weights.}
    \label{fig:dim_attention_heatmaps}
\end{figure*}
To verify whether the model’s predictions are grounded in genuine instructional evidence rather than spurious correlations, we examine the alignment between dimension-specific attention and human-annotated rationale spans. For each comment and each instructional facet, we compute the attention distribution from the corresponding dimension embedding to all tokens in the comment. The top-$k$ tokens with the highest attention scores, where $k$ is set to the top 20\% of tokens in the sentence, are treated as the model-predicted evidence for that facet. These tokens are then aligned with the manually annotated evidence spans, and we aggregate four evaluation metrics across the evaluation set, as summarized in Figure~\ref{fig:evidence_alignment}. Precision measures the proportion of model-highlighted tokens that fall inside the annotated evidence spans, indicating how selectively the model attends to true evidence. Recall measures the proportion of annotated evidence tokens that are captured by the top-attended tokens, reflecting the coverage of human rationales. IoU computes the intersection-over-union between the predicted and annotated evidence sets, providing a joint measure of coverage and selectivity. Finally, the entropy of the normalized attention distribution quantifies how concentrated the attention is over the token sequence, with lower entropy corresponding to more focused evidence attribution.

The curves in Figure~\ref{fig:evidence_alignment} show the evolution of these quantities over training epochs for all five instructional facets. In panel~(a), precision exhibits a clear upward trend across epochs: for most facets it increases from around 0.5--0.6 at early epochs to approximately 0.8 or higher by the end of training. This indicates that the tokens receiving the highest attention mass increasingly coincide with the annotated rationale tokens, suggesting that the model becomes more selective in highlighting truly evidential content.  Panel~(b) shows that recall decreases slightly as training progresses, though it remains relatively high overall. This mild decline suggests that, although the model continues to capture the majority of annotated rationale tokens, it gradually excludes a portion of spans that were originally annotated but contribute limited semantic value for prediction. In other words, as the model refines its internal notion of evidence, it becomes less likely to attend to tokens that appear in the annotations but do not function as strong predictive cues. This behaviour is consistent with the expectation that human-provided rationale spans may include peripheral or noisy content, and that the attention mechanism, through learning, prioritizes only the most informative parts of the evidence. Consistent with the trends observed in precision , panel~(c) reports a monotonic increase in IoU, moving from values around 0.5 at initialization to roughly 0.7--0.8 after convergence, which confirms that the overall overlap between predicted and gold evidence spans improves steadily as training proceeds. 

Panel~(d) presents the entropy of the attention distributions. For all facets, entropy gradually decreases over epochs, indicating that the attention assigned by each dimension embedding becomes less diffuse and more concentrated on a smaller subset of tokens. This reduction in entropy, combined with the simultaneous gains in precision, recall, and IoU, suggests that the model is not merely collapsing attention onto arbitrary positions, but is learning to allocate higher weights to tokens that correspond to semantically meaningful instructional evidence. Taken together, these results demonstrate that the learned dimension embeddings progressively acquire the ability to attend to the appropriate rationale spans in the comments, and that the decision process underlying facet-level predictions is increasingly supported by localized, human-interpretable evidence in the input text.

As a complementary qualitative analysis, Figure~\ref{fig:dim_attention_heatmaps} visualizes the token-level attention distributions induced by the five dimension embeddings at an early and a late training epoch. In the early stage, attention is relatively diffuse: for most dimensions, substantial mass is spread across a wide range of tokens, indicating that the model has not yet learned to distinguish facet-relevant evidence from surrounding context. By the final epoch, the heatmaps become markedly more concentrated, with sharper high-intensity bands emerging on a limited subset of tokens while large portions of the sentence receive only low attention. This shift is particularly evident for dimensions such as \emph{Quality}, where the initially near-uniform pattern evolves into a sparse structure that emphasizes only a few salient positions. Together with the quantitative trends in precision, IoU, and entropy, these visualizations support the conclusion that the learned dimension embeddings progressively acquire the ability to focus on facet-relevant evidence rather than distributing attention indiscriminately over the entire comment.

\section{Conclusion}
This paper introduced TeachScope, a pedagogically grounded benchmark for multi-dimensional instructional assessment, and TeachPro, a multi-facet teaching profiler tailored to its structure. Departing from traditional score-centric SET instruments and coarse comment-level polarity models, we formulated facet-level teaching evaluation as a multi-dimensional multi-class classification problem with explicit rationale spans, enabling both quantitative comparability and qualitative interpretability. TeachPro integrates a Cross-View Graph Synergy Network (DualGCN) for comment encoding, a Dimension-Anchored Evidence Encoder that uses facet-specific query embeddings to retrieve dimension-relevant rationales, and a unified low-rank perturbation classification head that shares parameters across facets while preserving dimension-specific decision boundaries. Extensive experiments on TeachScope demonstrate that TeachPro consistently and substantially outperforms a broad spectrum of recurrent, convolutional, and Transformer-based baselines in accuracy, F1, and agreement metrics, while ablation studies confirm the complementary contributions of each architectural component. Layer-wise representation analysis and evidence-alignment experiments further show that the model progressively disentangles facet representations and grounds its predictions in human-annotated rationale spans rather than in diffuse global impressions. In future work, we plan to extend TeachScope to additional institutions and languages, incorporate temporal and fairness-aware analyses of teaching trajectories, and explore human-in-the-loop deployment scenarios in which TeachPro supports, rather than replaces, expert-led instructional review and pedagogical development.

\printcredits

\bibliographystyle{model1-num-names}
\bibliography{cas-refs}

\end{document}